\documentclass[10pt,twocolumn,letterpaper]{article}

\usepackage{cvpr}
\usepackage{times}
\usepackage{epsfig}
\usepackage{graphicx}
\usepackage{amsmath}
\usepackage{amssymb}
\usepackage{balance}

\usepackage[pagebackref=true,breaklinks=true,letterpaper=true,colorlinks,bookmarks=false]{hyperref}

 \cvprfinalcopy 

\def\cvprPaperID{3303} 

\ifcvprfinal\fi

 \newif\ifarxiv
 \arxivtrue
\begin{document}

\title{3D Bounding Box Estimation Using Deep Learning and Geometry}

\author{Arsalan Mousavian\thanks{Work done as an intern at Zoox, Inc.}\\
George Mason University\\
{\tt\small amousavi@gmu.edu}
\and
Dragomir Anguelov\\
Zoox, Inc.\\
{\tt\small drago@zoox.com}
\and
John Flynn\\
Zoox, Inc.\\
{\tt\small john.flynn@zoox.com}
\and
Jana  Ko\v{s}eck\'a\\
George Mason University\\
{\tt\small kosecka@gmu.edu}
}

\maketitle
\definecolor{john_color}{rgb}{0,0.5,1}
\definecolor{arsalan_color}{rgb}{1,0,0}
\definecolor{jana_color}{rgb}{0.2,0.2,1}
\definecolor{drago_color}{rgb}{1,0.6,0}
\iftrue
\newcommand{\john}[1]{\textsf{\textcolor{john_color}{[{\bf John }: #1]}}}
\newcommand{\arsalan}[1]{\textsf{\textcolor{arsalan_color}{[{\bf Arsalan}: #1]}}}
\newcommand{\jana}[1]{\textsf{\textcolor{jana_color}{[{\bf Jana}: #1]}}}
\newcommand{\drago}[1]{\textsf{\textcolor{drago_color}{[{\bf Drago}: #1]}}}
\else
\newcommand{\john}[1]{}
\newcommand{\jana}[1]{}
\newcommand{\arsalan}[1]{}
\newcommand{\drago}[1]{}
\fi

\begin{abstract}
We present a method for 3D object detection and pose estimation from a single image. In contrast to current techniques that only regress the 3D orientation of an object, our method first regresses relatively stable 3D object properties using a deep convolutional neural network and then combines these estimates with geometric constraints provided by a 2D object bounding box to produce a complete 3D bounding box. The first network output estimates the 3D object orientation using a novel hybrid discrete-continuous loss, which significantly outperforms the L2 loss. The second output regresses the 3D object dimensions, which have relatively little variance compared to alternatives and can often be predicted for many object types. These estimates, combined with the geometric constraints on translation imposed by the 2D bounding box, enable us to recover a stable and accurate 3D object pose. We evaluate our method on the challenging KITTI object detection benchmark~\cite{KITTICVPR12} both on the official metric of 3D orientation estimation and also on the accuracy of the obtained 3D bounding boxes. Although conceptually simple, our method outperforms more complex and computationally expensive approaches that leverage semantic segmentation, instance level segmentation and flat ground priors~\cite{ChenUrtasunCVPR16} and  sub-category detection~\cite{xiang_cvpr15}\cite{xiang2016subcategory}. Our discrete-continuous loss also produces state of the art results for 3D viewpoint estimation on the Pascal 3D+ dataset\cite{XiangSavareseWACV14}.

\end{abstract}

\section{Introduction}

The problem of 3D object detection is of particular importance in robotic applications that require decision making or interactions with objects in the real world. 3D object detection recovers both the 6 DoF pose and the dimensions of an object from an image. While recently developed 2D detection algorithms are capable of handling large variations in viewpoint and clutter, accurate 3D object detection largely remains an open problem despite some promising recent work. The existing efforts to integrate pose estimation with state-of-the-art object detectors focus mostly on viewpoint estimation. They
exploit the observation that the appearance of objects changes as a function of viewpoint and that discretization of viewpoints (parametrized by azimuth and elevation) gives rise  to sub-categories which can be trained discriminatively~\cite{xiang_cvpr15}. 
In more restrictive driving scenarios alternatives to full 3D pose estimation explore  exhaustive sampling and scoring of all hypotheses~\cite{ChenUrtasunCVPR16} using a variety of contextual and semantic cues.



\begin{figure}
\begin{tabular}{c@{\hspace{1mm}}c}
\includegraphics[width=0.23\textwidth]{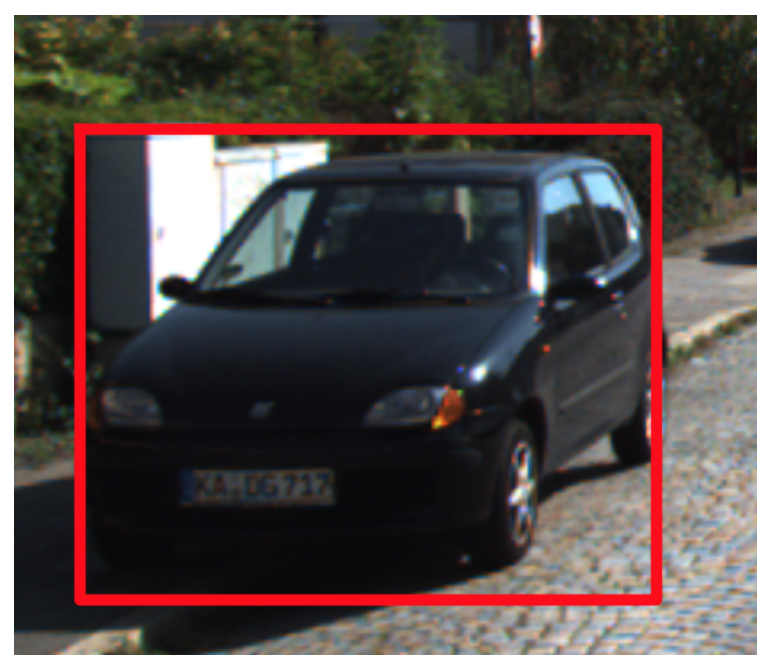} & \includegraphics[width=0.23\textwidth]{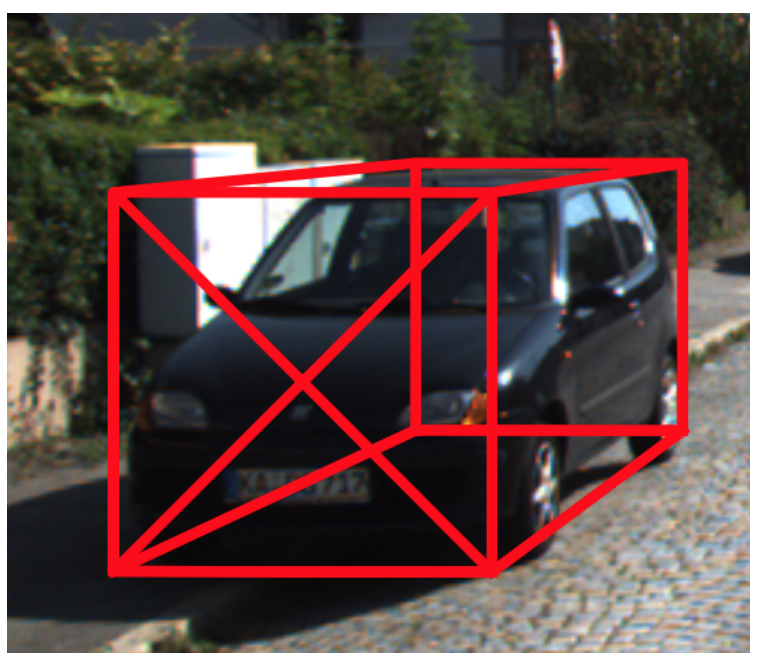}
\end{tabular}
\label{fig:overview}
\caption{Our method takes the 2D detection bounding box and estimates a 3D bounding box.}
\end{figure}

In this work, we propose a method that estimates the pose $(R, T) \in SE(3)$ and the dimensions of an object's 3D bounding box from a 2D bounding box and the surrounding image pixels. Our simple and efficient method is suitable for many real world applications including self-driving vehicles. The main contribution of our approach is in the choice of the regression parameters and the associated objective functions for the problem.  We first regress the orientation and object dimensions before combining these estimates with geometric constraints to produce a final 3D pose.  This is in contrast to previous techniques that attempt to directly regress to pose.
 
A state of the art 2D object detector~\cite{MSCNN2016} is extended by training a deep convolutional neural network (CNN) to regress the orientation of the object's 3D bounding box and its dimensions. Given estimated orientation and dimensions and the constraint that the projection of the 3D bounding box fits tightly into the 2D detection window, we recover the translation and the object's 3D bounding box. Although conceptually simple, our method is based on several important insights. We show that a novel \emph{MultiBin} discrete-continuous formulation of the orientation regression significantly outperforms a more traditional L2 loss.
Further constraining the 3D box by regressing to vehicle dimensions proves especially effective, since they are relatively low-variance and result in stable final 3D box estimates. 

We evaluate our method on the KITTI  \cite{KITTICVPR12} and Pascal 3D+\cite{XiangSavareseWACV14} datasets. On the KITTI dataset, we perform an in-depth comparison of our estimated 3D boxes to the results of other state-of-the-art 3D object detection algorithms
  \cite{xiang2016subcategory,ChenUrtasunCVPR16}. The official KITTI benchmark for 3D bounding box estimation only evaluates the 3D box orientation estimate. We introduce three additional performance metrics measuring the 3D box accuracy: distance to center of box, distance to the center of the closest bounding box face, and the overall bounding box overlap with the ground truth box, measured using 3D Intersection over Union (3D IoU) score. We demonstrate that given sufficient training data, our method is superior to the state of the art on all the above 3D metrics. Since the Pascal 3D+ dataset does not have the physical dimensions annotated and the intrinsic camera parameters are approximate, we only evaluate  viewpoint estimation accuracy showing that our \emph{MultiBin} module achieves state-of-the-art results there as well.

In summary, the main contributions of our paper include: 
1) A method to estimate an object's full 3D pose and dimensions from a 2D bounding box using the constraints provided by projective geometry and estimates of the object's orientation and size regressed using a deep CNN. In contrast to other methods, our approach does not require any preprocessing stages or 3D object models.
2) A novel discrete-continuous CNN architecture called \emph{MultiBin} regression for estimation of the object's orientation. 
3) Three new metrics for evaluating 3D boxes beyond their orientation accuracy for the KITTI dataset. 
4) An experimental evaluation demonstrating the effectiveness of our approach for KITTI cars, which also illustrates the importance of the specific choice of regression parameters within our 3D pose estimation framework.
5) Viewpoint evaluation on the Pascal 3D+ dataset.

\section{Related Work}
The classical problem of 6 DoF pose estimation of an object instance from a single 2D image has been considered previously as a purely geometric problem known as the \emph{perspective n-point problem (PnP)}. Several closed form and iterative solutions assuming correspondences between 2D keypoints in the image and a 3D model of the object can be found in~\cite{Lepetit09} and references therein. Other methods focus on constructing 3D models of the object instances and then finding the 3D pose in the image that best matches the model~\cite{Rothganger06,FerrariIJCV06}.

With the introduction of new challenging datasets \cite{KITTICVPR12,XiangSavareseWACV14,xiang2016objectnet3d,matzenICCV13}, 3D pose estimation has been extended to object categories, which requires handling both the appearance variations due to pose changes and the appearance variations within the category \cite{shapesKarTCM15,PepikCVPR15}. 
In ~\cite{PepikSchieleCVPR12, XiangSavareseWACV14} 
 the object detection framework of discriminative part based models (DPMs) is used to tackle the problem of pose estimation formulated jointly as a structured prediction problem, where each mixture component represents a different azimuth section. However, such approaches predict only an Euler angle subset with respect to the canonical object frame, while object dimensions and position are not estimated. 
 
An alternative direction is to exploit the availability of 3D shape models and use those for 3D hypothesis sampling and refinement. For example,  Mottaghi~\etal~\cite{mottaghi_cvpr15} sample the object viewpoint, position and size and then measure the similarity between rendered 3D CAD models of the object
and the detection window using HOG features. A similar method for estimating the pose using the projection of CAD model object instances has been explored by~\cite{ZhuDaniilidisICRA13} in a robotics table-top setting where the detection problem is less challenging. Given the coarse pose estimate obtained from a DPM-based detector, the continuous 6 DoF pose is refined by estimating the correspondences between the projected 3D model and the image contours. The evaluation was carried out on PASCAL3D+ or simple table top settings with limited clutter or scale variations. An extension of these methods to more challenging scenarios with significant occlusion has been explored in~\cite{3DVP15}, which uses dictionaries of 3D voxel patterns learned from 3D CAD models that characterize both the object's shape and commonly encountered occlusion patterns.

Recently, deep convolutional neural networks (CNN) have dramatically improved the performance of 2D object detection and several extensions have been proposed to include 3D pose estimation. In~\cite{TulsianiCVPR15} R-CNN~\cite{R-CNN_CVPR15} is used to detect objects and the resulting detected regions are passed as input to a pose estimation network. The pose network is initialized with VGG~\cite{Simonyan14c} and fine-tuned for pose estimation using ground truth annotations from Pascal 3D+. This approach is similar to~\cite{SuICCV15}, with the distinction of using separate pose weights for each category and a large number of synthetic images with pose annotation ground truth for training.
In~\cite{PoirsonBerg3DV16}, Poirson~\etal discretize the object viewpoint and train a deep convolutional network to jointly perform viewpoint estimation and 2D detection. The network shares the pose parameter weights across all classes. 
In~\cite{TulsianiCVPR15}, Tulsiani~\etal explore the relationship between coarse viewpoint estimation, followed by keypoint detection, localization and pose estimation. Pavlakos et al \cite{PavlakosICRA17}, used CNN to localize the keypoints and they used the keypoints and their 3D coordinates from meshes to recover the pose. However, their approach required training data with annotated keypoints. 



Several recent methods have explored 3D bounding box detection for driving scenarios and are most closely related to our method. Xiang~\etal~\cite{xiang_cvpr15,xiang2016subcategory} cluster the set of possible object poses into viewpoint-dependent subcategories. These subcategories are obtained by clustering 3D voxel patterns introduced  
previously~\cite{3DVP15}; 3D CAD models are required to learn the pattern dictionaries.
The subcategories capture both shape, viewpoint and occlusion patterns and are subsequently classified discriminatively~\cite{xiang2016subcategory} using deep CNNs. Another related approach by Chen~\etal~\cite{ChenUrtasunCVPR16}
addresses the problem by sampling 3D boxes
in the physical world assuming the flat ground plane constraint. The boxes are scored using high level contextual, shape and category specific features. All of the above approaches require complicated preprocessing including high level features such as segmentation or 3D shape repositories and may not be suitable for robots with limited computational resources. 

\section{3D Bounding Box Estimation}
In order to leverage the success of existing work on 2D object detection for 3D bounding box estimation, we use the fact that the perspective projection of a 3D bounding box should fit tightly within its 2D detection window. We assume that the 2D object detector has been trained to produce boxes that correspond to the bounding box of the projected 3D box. The 3D bounding box is described by its center $T = [t_x, t_y, t_z]^T$, dimensions $D=[d_x, d_y, d_z]$, and orientation $R(\theta, \phi, \alpha)$ , here paramaterized by the azimuth, elevation and roll angles. Given the pose of the object in the camera coordinate frame $(R, T) \in SE(3)$ and the camera intrinsics matrix $K$, the projection of a 3D point ${\bf X}_o = [X, Y, Z, 1]^T$ in the object's coordinate frame into the image ${\bf x} = [x,y,1]^T$  is:
\begin{equation}
\label{eq:projection}
{\bf x} = K \left[ \begin{matrix}
R & T \\
\end{matrix} \right]  {\bf X}_o
\end{equation}
Assuming that the origin of the object coordinate frame is at the center of the 3D bounding box and the object dimensions $D$ are known, the coordinates of the 3D bounding box vertices can be described simply by ${\bf X}_1 = [d_x/2, d_y/2, d_z/2]^T$, ${\bf X}_2 = [-d_x/2, d_y/2, d_z/2]^T$, $\hdots$ , ${\bf X}_8 =  [-d_x/2, -d_y/2, -d_z/2]^T$. 
The constraint that the 3D bounding box fits tightly into 2D detection window requires that each side of the 2D bounding box to be touched by the projection of at least one of the 3D box corners.
For example, consider the projection of one 3D corner ${\bf X_0}  = [d_x/2, -d_y/2, d_z/2]^T$ that touches the left side of the 2D bounding box with coordinate $x_{\mathit{min}}$.  This point-to-side correspondence constraint results in the equation: 
\begin{equation}
x_{\mathit{min}} = \left(K \left[ \begin{matrix}
R & T \\
\end{matrix} \right]  \left[ \begin{matrix}
d_x/2 \\ -d_y/2 \\ d_z/2 \\ 1
\end{matrix}\right] \right)_x
\label{eq:x}
\end{equation}
where $(.)_x$ refers to the $x$ coordinate from the perspective projection. Similar equations can be derived for the remaining 2D box side parameters $x_{\mathit{max}},y_{\mathit{min}}, y_{\mathit{max}}$. In total the sides of the 2D bounding box provide four constraints on the 3D bounding box. This is not enough to constrain the nine degrees of freedom (DoF) (three for translation, three for rotation, and three for box dimensions). 
There are several different geometric properties we could estimate from the visual appearance of the box to further constrain the 3D box. The main criteria is that they should be tied strongly to the visual appearance and further constrain the final 3D box. 

\subsection{Choice of Regression Parameters}
The first set of parameters that have a strong effect on the 3D bounding box is the orientation around each axis $(\theta, \phi, \alpha)$. Apart from them, we choose to regress the box dimensions $D$ rather than translation $T$ because the variance of the dimension estimate is typically smaller (e.g. cars tend to be roughly the same size) and does not vary as the object orientation changes: a desirable property if we are also regressing orientation parameters. Furthermore, the dimension estimate is strongly tied to the appearance of a particular object subcategory and is likely to be accurately recovered if we can classify that subcategory. 
\ifarxiv
In Sec.~\ref{sec:alternative_representation} we carried out experiments on regressing alternative parameters related to translation and found that choice of parameters matters: we obtained less accurate 3D box reconstructions using that parametrization. 
The CNN architecture and  the associated loss functions for  this regression problem are discussed in Sec.~\ref{sec:cnn_regression}.
\fi

\subsection{Correspondence Constraints}
\begin{figure}[t]
\centering
\begin{tabular}{cc}
\includegraphics[width=0.17\textwidth]{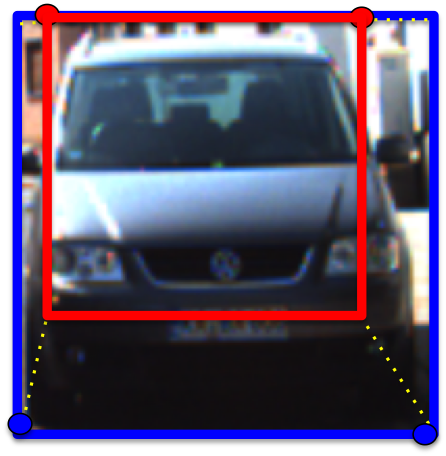} & \includegraphics[width=0.17\textwidth]{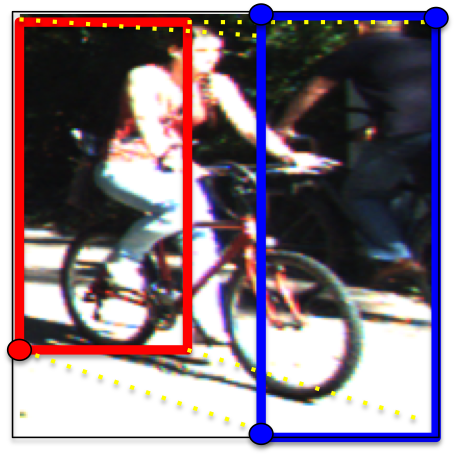} \\
\includegraphics[width=0.17\textwidth]{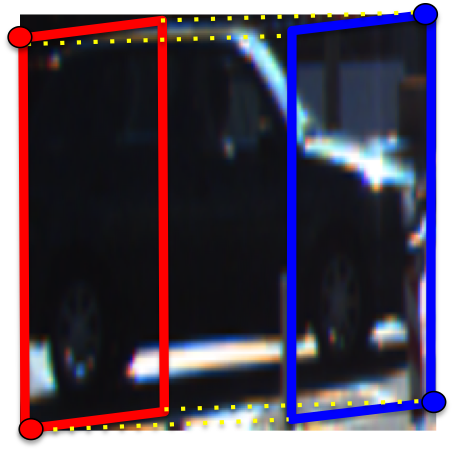} & \includegraphics[width=0.17\textwidth]{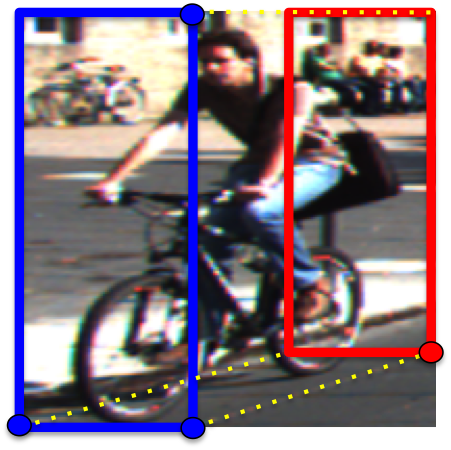}
\end{tabular}
\caption{Correspondence between the 3D box and 2D bounding box: Each figure shows a 3D bbox that surrounds an object. The front face is shown in blue and the rear face is in red. The 3D points that are active constraints in each of the images are shown with a circle (best viewed in color). }
\label{fig:cubes}
\end{figure}

Using the regressed dimensions and orientations of the 3D box by CNN and 2D detection box we can solve for the translation $T$ that minimizes the reprojection error with respect to the initial 2D detection box constraints in Equation~\ref{eq:x}. Details of how to solve for translation are included in the supplementary material \cite{Supplementary}. Each side of the 2D detection box can correspond to any of the eight corners of the 3D box which results in $8^4 = 4096$ configurations. Each different configuration involves solving an over-constrained system of linear equations which is computationally fast and can be done in parallel. In many scenarios the objects can be assumed to be always upright. In this case, the 2D box top and bottom correspond only to the projection of vertices from the top and bottom of the 3D box, respectively, which reduces the number of correspondences to 1024. Furthermore, when the  relative object roll is close to zero, the vertical 2D box side coordinates $x_{min}$ and $x_{max}$ can only correspond to projections of points from vertical 3D box sides.
Similarly, $y_{min}$ and $y_{max}$ can only correspond to point projections from the horizontal 3D box sides.
Consequently, each vertical side of the 2D detection box can correspond to $[\pm d_x/2, ., \pm d_z/2]$ and each horizontal side of the 2D bounding corresponds to $[., \pm d_y/2, \pm d_z/2]$, yielding $4^4 = 256$ possible configurations. In the KITTI dataset, object pitch and roll angles are both zero, which further reduces of the number of configurations to 64.  Fig.~\ref{fig:cubes} visualizes some of the possible correspondences between 2D box sides and 3D box points that can occur.

\section{CNN Regression of 3D Box Parameters}
\label{sec:cnn_regression}
In this section, we describe our approach for regressing the 3D bounding box orientation and dimensions.
\subsection{MultiBin Orientation Estimation}
\label{subsec:orientation_regression}
\begin{figure}
\begin{tabular}{c}
\includegraphics[width=0.4\textwidth]{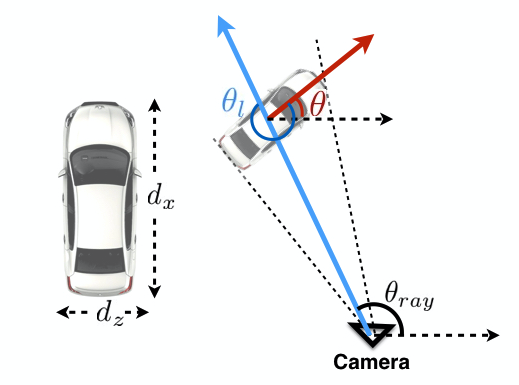}
\end{tabular}
\caption{Left: Car dimensions, the height of the car equals $d_y$. Right: Illustration of local orientation $\theta_l$, and global orientation of a car $\theta$. The local orientation is computed with respect to the ray that goes through the center of the crop. The center ray of the crop is indicated by the blue arrow. Note that the center of crop may not go through the actual center of the object. Orientation of the car $\theta$ is equal to $\theta_{ray} + \theta_l$ . The network is trained to estimate the local orientation $\theta_l$.}
\label{fig:ray}
\end{figure}
\begin{figure}
\includegraphics[width=0.45\textwidth]{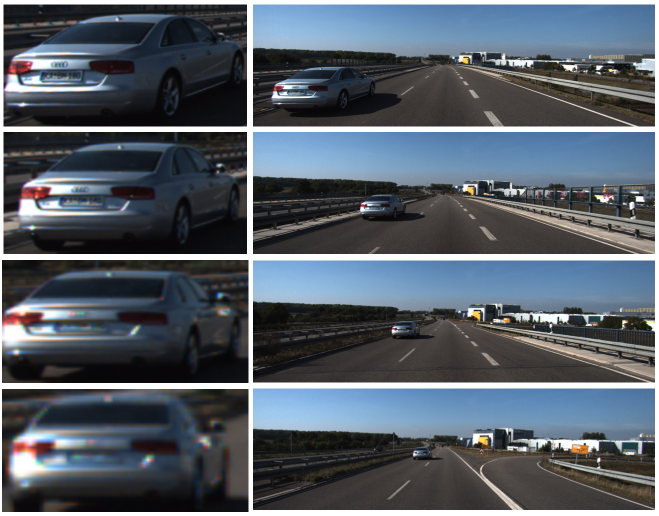}
\caption{Left: cropped image of a car passing by. Right: Image of whole scene. As it is shown the car in the cropped images rotates while the car direction is constant among all different rows. }
\label{fig:car_passing}
\end{figure}

Estimating the global object orientation $R \in SO(3)$ in the camera reference frame from only the contents of the detection window crop is not possible, as the location of the crop within the image plane is also required.  Consider the rotation $R(\theta)$ parametrized  only  by azimuth $\theta$ (yaw). Fig.~\ref{fig:car_passing} shows an example of a car moving in a straight line. Although the global orientation $R(\theta)$ of the car (its 3D bounding box) does not change, its local orientation $\theta_l$  with respect to the ray through the crop center does, and generates changes in the appearance of the cropped image. 

We thus regress to this local orientation  $\theta_l$.
Fig.~\ref{fig:car_passing} shows an example, where the local orientation angle $\theta_l$ and the ray angle change in such a way that their combined effect is a constant global orientation of the car. 
Given intrinsic camera parameters, the ray direction at a particular pixel is trivial to compute. At inference time we combine this ray direction at the crop center with the estimated local orientation in order to compute the global orientation of the object. 

It is known that using the L2 loss is not a good fit for many complex multi-modal regression problems. The L2 loss encourages the network to minimize to average loss across all modes, which results in an estimate that may be poor for any single mode. This has been observed in the context of the image colorization problem, where the L2 norm produces unrealistic average colors for items like clothing~\cite{ZhangColorization16}. Similarly, object detectors such as Faster R-CNN~\cite{YOLO_CVPR16} and SSD~\cite{SSD_ECCV16} do not regress the bounding boxes directly: instead they divide the space of the bounding boxes into several discrete modes called \emph{anchor boxes} and then estimate the continuous offsets that need to be applied to each anchor box. 

We use a similar idea in our proposed \emph{MultiBin} architecture for orientation estimation. We first discretize the orientation angle and divide it into $n$ overlapping bins. For each bin, the CNN network estimates both a confidence probability $c_i$ that the output angle lies inside the $i^{th}$ bin and the residual rotation correction that needs to be applied to the orientation of the center ray of that bin in order to obtain the output angle. The residual rotation is represented by two numbers, for the sine and the cosine of the angle. This results in 3 outputs for each bin $i$: $(c_i, \cos( \Delta \theta_i), \sin (\Delta \theta_i))$. Valid cosine and sine values are obtained by applying an L2 normalization layer on top of a 2-dimensional input. The total loss for the MultiBin orientation is thus:
\begin{equation}
\label{eq:multibin_loss}
L_{\theta} = L_{conf} + w \times L_{loc}
\end{equation}
\begin{figure}
\centering
\includegraphics[width=0.35\textwidth]{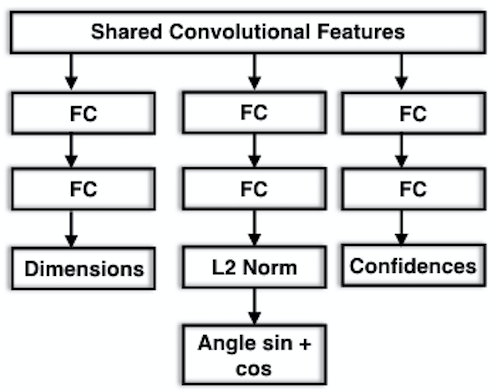}
\caption{Proposed architecture for MultiBin estimation for orientation and dimension estimation. It consists of three branches. The left branch is for estimation of dimensions of the object of interest. The other branches are for computing the confidence for each bin and also compute the $cos(\Delta \theta)$ and $sin(\Delta \theta)$ of each bin}
\label{fig:net}
\end{figure}
The confidence loss $L_{conf}$ is equal to the softmax loss of the confidences of each bin. $L_{loc}$ is the loss that tries to minimize the difference between the estimated angle and the ground truth angle in each of the bins that covers the ground truth angle, with adjacent bins having overlapping coverage. In the localization loss $L_{loc}$, all the bins that cover the ground truth angle are forced to estimate the correct angle. The localization loss tries to minimize the difference between the ground truth and all the bins that cover that value which is equivalent of maximizing cosine distance as it is shown in supplementary material \cite{Supplementary}. Localization loss $L_{loc}$ is computed as following:
\begin{equation}
\label{eq:localization_loss}
L_{loc} = -\frac{1}{n_{\theta^*}} \sum cos(\theta^* - c_i - \Delta \theta_i)
\end{equation}
where $n_{\theta^*}$ is the number of bins that cover ground truth angle $\theta ^*$, $c_i$ is the angle of the center of bin $i$ and $\Delta \theta_i$ is the change that needs to be applied to the center of bin $i$.

During inference, the bin with maximum confidence is selected and the final output is computed by applying the estimated  $\Delta \theta$ of that bin to the center of that bin. The {\em MultiBin} module has 2 branches. One for computing the confidences $c_i$ and the other for computing the cosine and sine of $\Delta \theta$. As a result, $3n$ parameters need to be estimated for $n$ bins.

In the KITTI dataset cars, vans, trucks, and buses are all different categories and the distribution of the object dimensions for category instances is low-variance and unimodal. For example, the dimension variance for cars and cyclists is on the order of several centimeters. Therefore, rather than using a discrete-continuous loss like the {\em MultiBin} loss above, we use directly the L2 loss. As is standard, for each dimension we estimate the residual relative to the mean parameter value computed over the training dataset. The loss for dimension estimation $L_{dims}$ is computed as follows:
\begin{equation}
\label{eq:dims_loss}
L_{dims} = \frac{1}{n} \sum{(D^* - \bar{D} - \delta)^2},
\end{equation}
where $D^*$ are the ground truth dimensions of the box, $\bar{D}$ are the mean dimensions for objects of a certain category and $\delta$ is the estimated residual with respect to the mean that the network predicts.

The CNN architecture of our parameter estimation module is shown in Figure~\ref{fig:net}. 
There are three branches: two branches for orientation estimation and one branch for dimension estimation. All of the branches are derived from the same shared convolutional features and the total loss is the weighted combination of $L = \alpha \times L_{dims} + L_{\theta}.$
\section{Experiments and Discussions}
\subsection{Implementation Details}
 We performed our experiments on the KITTI~\cite{KITTICVPR12} and Pascal 3D+\cite{XiangSavareseWACV14} datasets.

\begin{figure*}
\centering
\ifarxiv
\begin{tabular}{c@{\hspace{1mm}}c}
\includegraphics[width=0.4\textwidth]{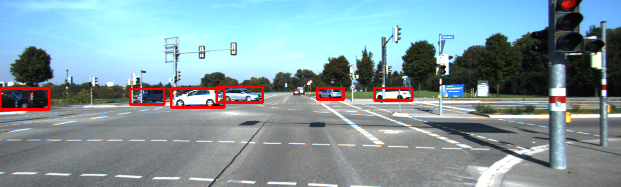}&
\includegraphics[width=0.4\textwidth]{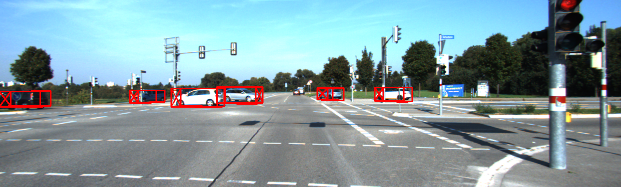}\\
\includegraphics[width=0.4\textwidth]{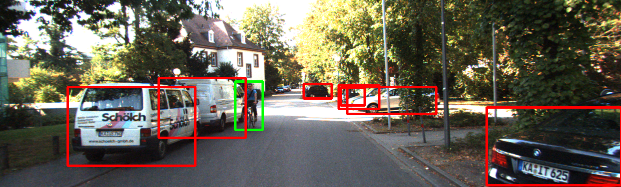}&
\includegraphics[width=0.4\textwidth]{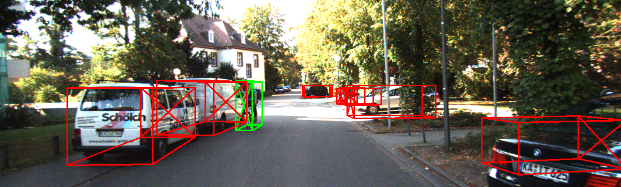}\\
\includegraphics[width=0.4\textwidth]{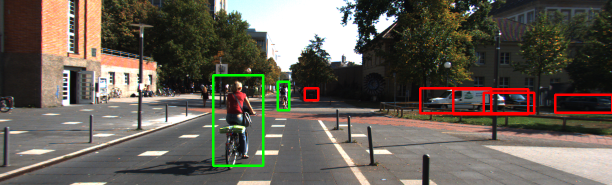}&
\includegraphics[width=0.4\textwidth]{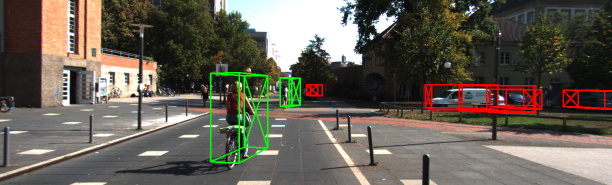}\\
\includegraphics[width=0.4\textwidth]{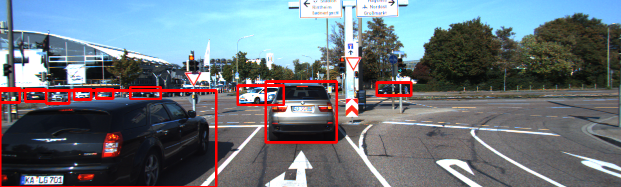}&
\includegraphics[width=0.4\textwidth]{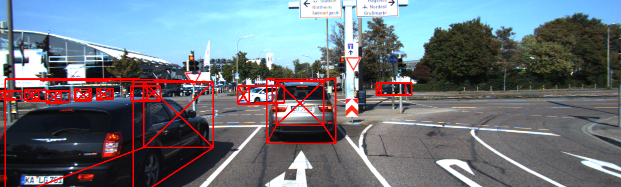}\\
\end{tabular}
\caption{Qualitative illustration of the 2D detection boxes and the estimated 3D projections, in red for cars and green for cyclists.}
\else
\begin{tabular}{c@{\hspace{1mm}}c}
\includegraphics[width=0.4\textwidth]{final_cr/000033.png}&
\includegraphics[width=0.4\textwidth]{final_cr/000164.png}\\
\includegraphics[width=0.4\textwidth]{final_cr/000234.png}&
\includegraphics[width=0.4\textwidth]{final_cr/001076.png}\\
\end{tabular}
\caption{Qualitative illustration of the estimated 3D projections, in red for cars and green for cyclists.}
\fi

\label{fig:qualitative}
\end{figure*}

 \noindent\textbf{KITTI dataset:} The KITTI dataset has a total of 7481 training images. We train the MS-CNN~\cite{MSCNN2016} object detector to produce 2D boxes and then estimate 3D boxes from 2D detection boxes whose scores exceed a threshold. 
For regressing 3D parameters, we use a pretrained VGG network~\cite{Simonyan14c} without its FC layers and add our 3D box module, which is shown in Fig.~\ref{fig:net}. In the module, the first FC layers in each of the orientation branches have 256 dimensions, while the first FC layer for dimension regression has a dimension of 512. During training, each ground truth crop is resized to 224x224. In order to make the network more robust to viewpoint changes and occlusions, the ground truth boxes are jittered and the ground truth $\theta_l$ is changed to account for the movement of the center ray of the crop. In addition, we added color distortions and also applied mirroring to images at random. The network is trained with SGD using a fixed learning rate of $0.0001$. The training is run for 20K iterations with a batch size of 8 and the best model is chosen by cross validation. Fig.~\ref{fig:qualitative} shows the qualitative visualization of estimated 3D boxes for cars and cyclists on our KITTI validation set. We used two different training/test splits for our experiments. The first split was used to report results on the official KITTI test set and uses the majority of the available training images. The second split is identical to the one used by  SubCNN~\cite{xiang2016subcategory}, in order to enable fair comparisons. They use half of the available data for validation. 

\noindent\textbf{Pascal3D+ dataset:} The dataset consists of images from Pascal VOC and Imagenet for 12 different categories that are annotated with 6 DoF pose. Images from the Pascal training set and Imagenet are used for training and the evaluation is done on the Pascal validation set. Unlike KITTI, the intrinsic parameters are approximate and therefore it is not possible to recover the true physical object dimensions. Therefore we only evaluate on 3 DoF viewpoint estimation to show the effectiveness of our \emph{MultiBin} loss.
We used $C \times 3$ \emph{MultiBin} modules to predict 3 angles for each of the $C$ classes. 
For a fair comparison with \cite{TulsianiCVPR15}, we kept the $fc6$ and $fc7$ layers of VGG and eliminated the separate convolution branches of our \emph{MultiBin} modules. All the necessary inputs are generated using a single fully connected layer that takes $fc7$ as input. We also reused the hyperparameters chosen in \cite{TulsianiCVPR15} for training our model.  

\ifarxiv
\begin{figure*}
\begin{tabular}{c@{\hspace{1mm}}c@{\hspace{1mm}}c@{\hspace{1mm}}c@{\hspace{1mm}}c@{\hspace{1mm}}c@{\hspace{1mm}}c@{\hspace{1mm}}c}
\includegraphics[width = 0.12\textwidth]{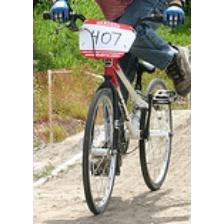}&
\includegraphics[width = 0.12\textwidth]{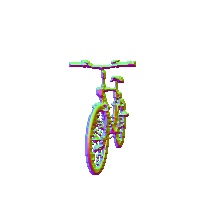}&
\includegraphics[width = 0.12\textwidth]{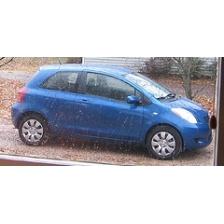}&
\includegraphics[width = 0.12\textwidth]{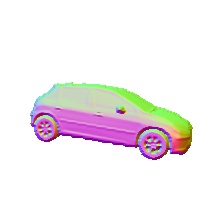}&
\includegraphics[width = 0.12\textwidth]{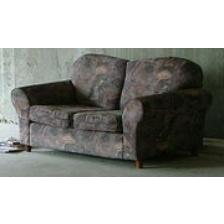}&
\includegraphics[width = 0.12\textwidth]{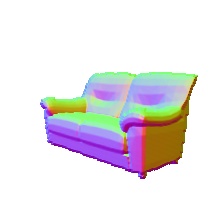}&
\includegraphics[width = 0.12\textwidth]{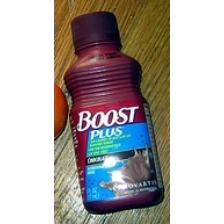}&
\includegraphics[width = 0.12\textwidth]{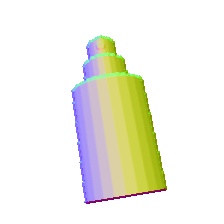}\\

\includegraphics[width = 0.12\textwidth]{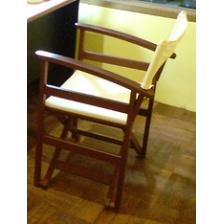}&
\includegraphics[width = 0.12\textwidth]{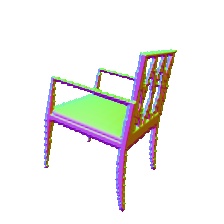}&
\includegraphics[width = 0.12\textwidth]{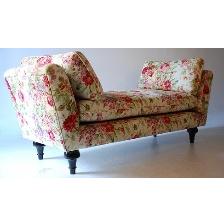}&
\includegraphics[width = 0.12\textwidth]{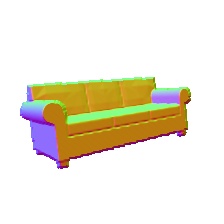}&
\includegraphics[width = 0.12\textwidth]{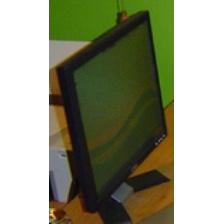}&
\includegraphics[width = 0.12\textwidth]{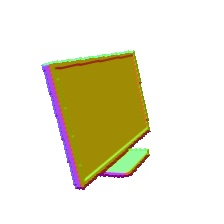}&
\includegraphics[width = 0.12\textwidth]{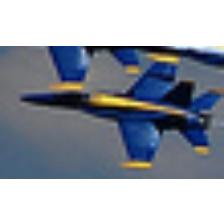}&
\includegraphics[width = 0.12\textwidth]{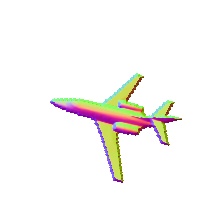}\\

\includegraphics[width = 0.12\textwidth]{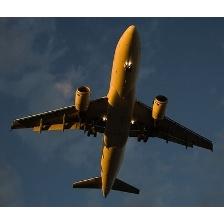}&
\includegraphics[width = 0.12\textwidth]{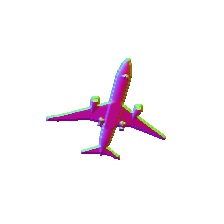}&
\includegraphics[width = 0.12\textwidth]{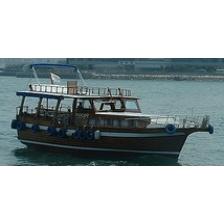}&
\includegraphics[width = 0.12\textwidth]{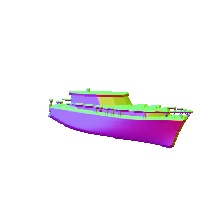}&
\includegraphics[width = 0.12\textwidth]{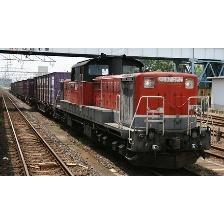}&
\includegraphics[width = 0.12\textwidth]{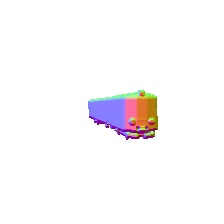}&
\includegraphics[width = 0.12\textwidth]{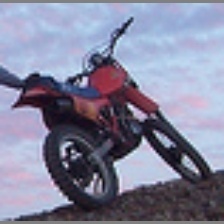}&
\includegraphics[width = 0.12\textwidth]{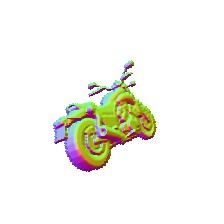}\\

\end{tabular}
\caption{Visualization of Estimated Poses on Pascal3D+ dataset}
\label{fig:pascal_mesh}
\end{figure*}
\else
\begin{figure}
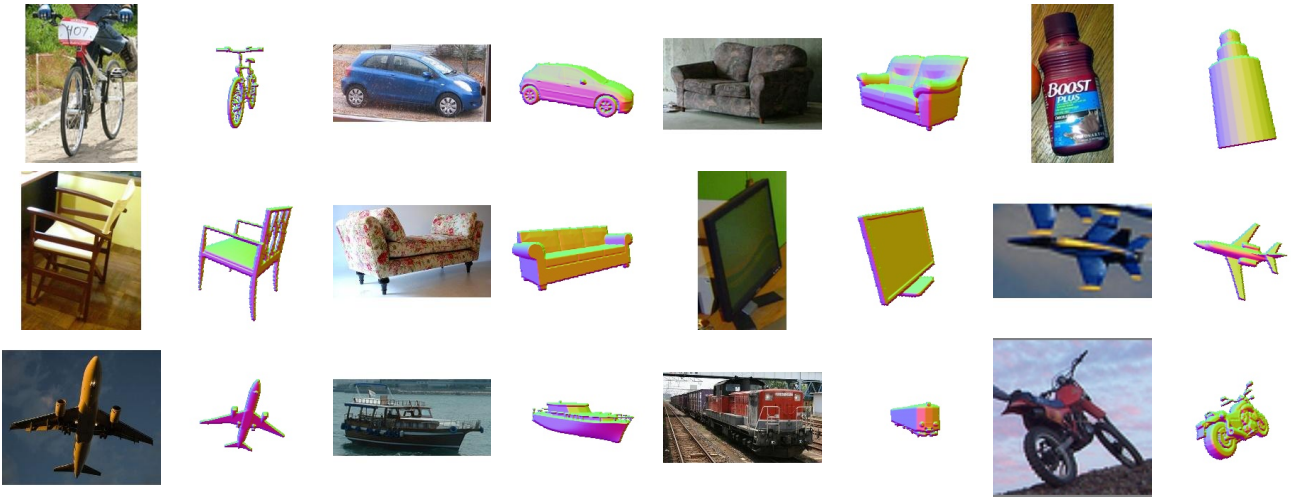

\begin{tabular}{c@{\hspace{1mm}}c@{\hspace{1mm}}c@{\hspace{1mm}}c}
\includegraphics[width = 0.12\textwidth]{pascal_mesh/71_a.jpg}&
\includegraphics[width = 0.12\textwidth]{pascal_mesh/71_b.jpg}&
\includegraphics[width = 0.12\textwidth]{pascal_mesh/3_a.jpg}&
\includegraphics[width = 0.12\textwidth]{pascal_mesh/3_b.jpg}\\
\includegraphics[width = 0.12\textwidth]{pascal_mesh/12_a.jpg}&
\includegraphics[width = 0.12\textwidth]{pascal_mesh/12_b.jpg}&
\includegraphics[width = 0.12\textwidth]{pascal_mesh/16_a.jpg}&
\includegraphics[width = 0.12\textwidth]{pascal_mesh/16_b.jpg}\\

\includegraphics[width = 0.12\textwidth]{pascal_mesh/22_a.jpg}&
\includegraphics[width = 0.12\textwidth]{pascal_mesh/22_b.jpg}&
\includegraphics[width = 0.12\textwidth]{pascal_mesh/30_a.jpg}&
\includegraphics[width = 0.12\textwidth]{pascal_mesh/30_b.jpg}\\
\includegraphics[width = 0.12\textwidth]{pascal_mesh/37_a.jpg}&
\includegraphics[width = 0.12\textwidth]{pascal_mesh/37_b.jpg}&
\includegraphics[width = 0.12\textwidth]{pascal_mesh/39_a.jpg}&
\includegraphics[width = 0.12\textwidth]{pascal_mesh/39_b.jpg}\\

\end{tabular}
\caption{Visualization of Estimated Poses on Pascal3D+ dataset}
\label{fig:pascal_mesh}
\end{figure}
\fi

\subsection{3D Bounding Box Evaluation}
\begin{table*}
\centering
\begin{tabular}{|c||ccc|ccc|ccc|}
\hline
Method & & Easy & & & Moderate & & & Hard & \\
& AOS & AP & OS & AOS & AP & OS & AOS & AP & OS\\
\hline
3DOP\cite{3DOPNIPS15} & 91.44\% & {\bf 93.04\%} & 0.9828&  86.10\% & 88.64\% & 0.9713 & 76.52\% & 79.10\% &  0.9673\\
Mono3D\cite{ChenUrtasunCVPR16} & 91.01\% & 92.33\% & 0.9857 & 0.8662\% & 88.66\% & 0.9769 & 76.84\% & 78.96\% & 0.9731 \\
SubCNN\cite{xiang2016subcategory} & 90.67\% & 90.81\%& 0.9984 & 88.62\% & 89.04\%& 0.9952  &  {\bf 78.68\%} & {\bf 79.27\%} & 0.9925 \\
Our Method &  {\bf 92.90\%} & 92.98\% & {\bf 0.9991} & {\bf 88.75\%} & {\bf 89.04\%} & {\bf 0.9967} &  76.76\% & 77.17\% & {\bf 0.9946} \\
\hline
\end{tabular}
\caption{Comparison of the Average Orientation Estimation (AOS), Average Precision (AP) and Orientation Score (OS) on official KITTI dataset for cars. Orientation score is the ratio between AOS and AP.}
\label{tab:kitti_car}
\end{table*}

\ifarxiv
\begin{table*}
\centering
\label{tab:kitti_cyclist}
\begin{tabular}{|c||ccc|ccc|ccc|}
\hline
Method & & Easy & & & Moderate & & & Hard & \\
& AOS & AP & OS & AOS & AP & OS & AOS & AP & OS\\
\hline
3DOP\cite{3DOPNIPS15} & 70.13\% & 78.39\% & 0.8946 & 58.68\% & 68.94\% & 0.8511 &  52.32\% & 61.37\% & 0.8523 \\
Mono3D\cite{ChenUrtasunCVPR16} & 65.56\% & 76.04\% & 0.8621 & 54.97\% & 66.36\% & 0.8283 & 48.77\% & 58.87\% & 0.8284 \\
SubCNN\cite{xiang2016subcategory} & 72.00\% & 79.48\% & 0.9058 & 63.65\% & 71.06\% & 0.8957 & 56.32\% & 62.68\% & 0.8985\\
Our Method & 69.16\% & 83.94\% & 0.8239&  59.87\% & 74.16\% & 0.8037 & 52.50\% & 64.84\% & 0.8096\\
\hline
\end{tabular}
\caption{AOS comparison on the official KITTI dataset for cyclists. Our purely data-driven model is not able to match the performance of methods that use additional features and assumptions with just 1100 training examples.}
\end{table*}
\fi
\noindent\textbf{KITTI orientation accuracy.}  The official 3D metric of the KITTI dataset is Average Orientation Similarity (AOS), which is defined in~\cite{KITTICVPR12} and multiplies the average precision (AP) of the 2D detector with the average cosine distance similarity for azimuth orientation. Hence, AP is by definition the upper bound of AOS.  
   At the time of publication, we are first among all methods in terms of AOS for easy car examples and first among all non-anonymous methods for moderate car examples on the KITTI leaderboard. Our results are summarized in Table~\ref{tab:kitti_car}, which shows that we outperform all the recently published methods on orientation estimation for cars. For moderate cars we outperform SubCNN~\cite{xiang2016subcategory} despite having similar AP, while for hard examples we outperform 3DOP~\cite{3DOPNIPS15} despite much lower AP. The ratio of AOS over AP for each method is representative of how each method performs only on orientation estimation, while factoring out the 2D detector performance. We refer to this score as Orientation Score (OS), which represents the error $(1 + cos(\Delta \theta))/2$ averaged across all examples. OS can be converted back to angle error by the $acos(2 * OS - 1)$ formula, resulting in $3^\circ$ error for easy, $6^\circ$ for moderate, and $8^\circ$ on hard cars for our MultiBin model on the official KITTI test set. Our method is the only one that does not rely on computing additional features such as stereo, semantic segmentation, instance segmentation and does not need preprocessing as in ~\cite{xiang2016subcategory} and~\cite{xiang_cvpr15}. 

\noindent\textbf{Pascal3D+ viewpoint accuracy.} Two metrics are used for viewpoint accuracy: Median Error $MedErr$ and the percentage of the estimations that are within $\frac{\pi}{6}$ of the groundtruth viewpoint $\mathit{Acc}_{\frac{\pi}{6}}$. The distance between rotations is computed as $\Delta (R_1, R_2) = \frac{||\log (R_1^T R_2)||_F}{\sqrt{2}}$. The evaluation is done using the groundtruth bounding boxes. Table \ref{tab:pascal_viewpoint} shows that \emph{MultiBin} modules are more effective than discretized classification \cite{TulsianiCVPR15} and also keypoint based method of \cite{PavlakosICRA17} which is based on localizing keypoints and solving a sophisticated optimization to recover the pose.

\noindent\textbf{MultiBin loss analysis.} Table~\ref{tab:bins} shows the effect of choosing a different number of bins for the Multibox loss on both KITTI and Pascal3D+. In both datasets, using more than one bin consistently outperforms the single-bin variant, which is equivalent to the L2 loss. On KITTI, the best performance is achieved with 2 bins while 8 bins works the best for Pascal3D+. This is due to the fact that the viewpoint distribution in the Pascal3D+ dataset is more diverse. As Table \ref{tab:bins} shows, over-binning eventually decreases the effectiveness of the method, as it decreases the training data amount for each bin. We also experimented with different widths of the fully connected layers (see Table \ref{tab:tab1}) and found that increasing the width of the FC layers further yielded some limited gains even beyond width 256.

\begin{figure*}
\centering
\begin{tabular}{c@{\hspace{1mm}}c@{\hspace{1mm}}c}
\includegraphics[width=0.3\textwidth]{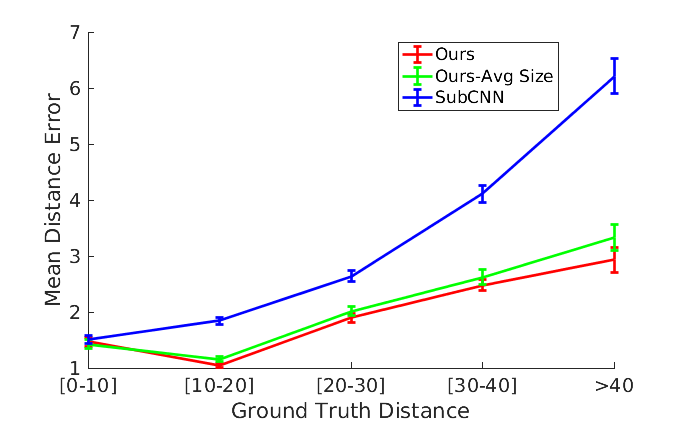}&
\includegraphics[width=0.3\textwidth]{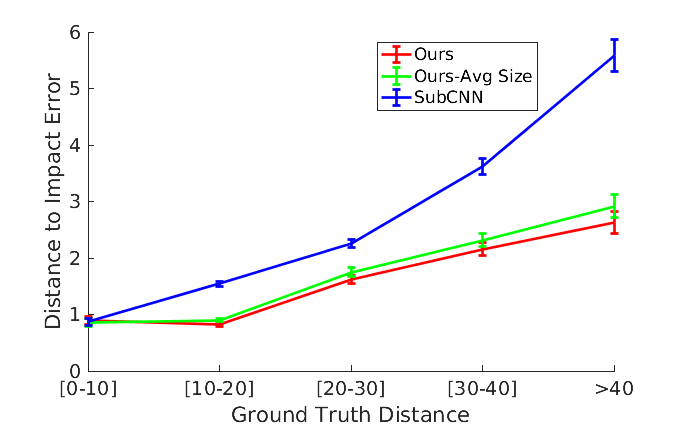}&
\includegraphics[width=0.3\textwidth]{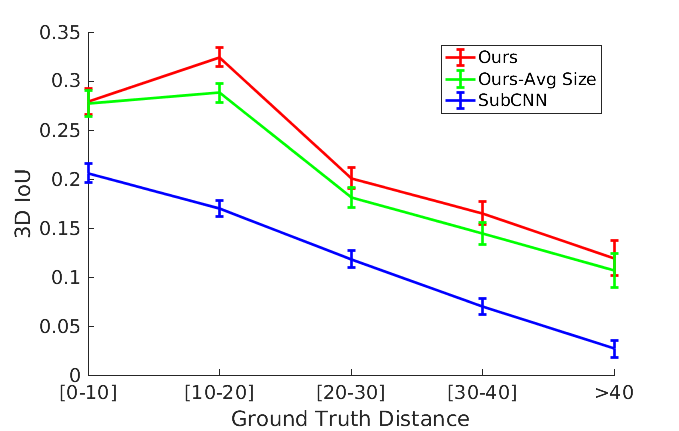} \\
\end{tabular}
\caption{3D box metrics for KITTI cars. Left: Mean distance error for box center, in meters. Middle: Error in estimating the closest distance from the 3D box to the camera, which is proportional to time-to-impact for driving scenarios. Right: 3D IoU between the predicted and ground truth 3D bounding boxes.}
\label{fig:metrics}
\end{figure*}

\noindent\textbf{3D bounding box metrics and comparison.}
The orientation estimation loss evaluates only a subset of 3D bounding box parameters. To evaluate the accuracy of the rest, we introduce 3 metrics, on which we compare our method against SubCNN~\cite{xiang2016subcategory} for KITTI cars.
The first metric is the average error in estimating the 3D coordinate of the center of the objects. The second metric is the average error in estimating the closest point of the 3D box to the camera. This metric is important for driving scenarios where the system needs to avoid hitting obstacles. The last metric is the 3D intersection over union (3D IoU) which is the ultimate metric utilizing all parameters of the estimated 3D bounding boxes. In order to factor away the 2D detector performance for a side-by-side comparison, we kept only the detections from both methods where the detected 2D boxes have $\textrm{IoU} \geq 0.7$. As Fig.~\ref{fig:metrics} shows, our method outperforms the SubCNN method~\cite{xiang2016subcategory}, the current state of the art, across the board in all 3 metrics. Despite this, the 3D IoU numbers are significantly smaller than those that 2D detectors typically obtain on the corresponding 2D metric. This is due to the fact that 3D estimation is a more challenging task, especially as the distance to the object increases. For example, if the car is 50m away from the camera, a translation error of 2m corresponds to about half the car length. Our method handles increasing distance well, as its error for the box center and closest point metrics in Fig.~\ref{fig:metrics} increases approximately linearly with distance, compared to SubCNN's super-linear degradation. To evaluate the importance of estimating the car dimensions, we evaluated a variant of our method that uses average sizes instead of estimating them. The evaluation shows that regressing the dimensions makes a difference in all the 3D metrics.
To facilitate comparisons with future work on this problem, we have made the estimated 3D boxes on the split of \cite{3DVP15} available at \url{http://bit.ly/2oaiBgi}.

\noindent\textbf{Training data requirements.}
One downside of our method is that it needs to learn the parameters for the fully connected layers; it requires more training data than methods that use additional information. To verify this hypothesis, we repeated the experiments for cars but limited the number of training instances to 1100. The same method that achieves $0.9808$ in Table~\ref{tab:bins} with $10828$ instances can only achieve $0.9026$ on the same test set. Moreover, our results on the official KITTI set is significantly better than the split of \cite{3DVP15} (see Table~\ref{tab:kitti_car}) because yet more training data is used for training. A similar phenomenon is happening for the KITTI cyclist task. The number of cyclist instances are much less than the number of car instances (1144 labeled cyclists vs 18470 labeled cars). As a result, there is not enough training data for learning the parameters of the fully connected layer well. Although our purely data-driven method achieves competitive results on the cyclists
\ifarxiv 
(see Table~\ref{tab:kitti_cyclist})
\fi
, it cannot outperform other methods that use additional features and assumptions.


\begin{table*}
\centering
\begin{tabular}{c | c c c c c c c c c c c c| c }
\hline
& aero &bike& boat& bottle& bus& car& chair& table& mbike& sofa& train& tv & mean\\
\hline
$\mathit{MedErr}$(\cite{TulsianiCVPR15}) & 13.8 &17.7& {\bf 21.3}& 12.9& 5.8& 9.1& 14.8& 15.2& 14.7& 13.7& 8.7& 15.4  & 13.6\\
$\mathit{MedErr}$(\cite{PavlakosICRA17}) & {\bf 8.0}& 13.4& 40.7& 11.7& {\bf 2.0}& {\bf 5.5} & {\bf 10.4} & N/A & N/A & {\bf 9.6} & 8.3 & 32.9 & N/A\\
$\mathit{MedErr}$(Ours) &  13.6 & {\bf 12.5} & 22.8 & {\bf 8.3} & 3.1&  5.8 & 11.9 & {\bf 12.5} & {\bf 12.3} & 12.8 &{\bf 6.3}& {\bf 11.9} & {\bf 11.1}\\
\hline
$\mathit{Acc}_\frac{\pi}{6}$(\cite{TulsianiCVPR15})& {\bf 0.81}& 0.77& {\bf 0.59}& {\bf 0.93}& {\bf 0.98}& 0.89& {\bf 0.80}& 0.62& {\bf 0.88}& {\bf 0.82}& 0.80& 0.80& 0.8075\\
$\mathit{Acc}_\frac{\pi}{6}$(Ours)& 0.78& {\bf 0.83}& 0.57& {\bf 0.93} &0.94& {\bf 0.90}& {\bf 0.80}& {\bf 0.68}& 0.86& {\bf 0.82}& {\bf 0.82}& {\bf 0.85} & {\bf 0.8103} 
\end{tabular}
\caption{ Viewpoint Estimation with Ground Truth box on Pascal3D+}
\label{tab:pascal_viewpoint}
\end{table*}

\begin{table}
\scalebox{0.85}{
\begin{tabular}{|c c||c|c|c|c|c|}
\hline
dataset & \# of Bins & 1 & 2 & 4 & 8 & 16 \\
\hline
KITTI & OS & 0.89 & {\bf 0.98} & 0.97 &  0.97 & 0.96\\
Pascal3D+ & $\mathit{Acc}_\frac{\pi}{6}$ & 0.65 & 0.72 & 0.78 & {\bf 0.81} & 0.77 \\
\hline
\end{tabular}
}
\caption{The effect of the number of bins on viewpoint estimation in KITTI and Pascal3D+ datasets}
\label{tab:bins}
\end{table}

\begin{table}
\centering
\scalebox{0.8}{
\begin{tabular}{|c|ccccc|}
\hline
FC & 64 & 128 & 256 & 512 & 1024\\
\hline
OS & 0.9583 & 0.9607 & 0.9808 & 0.9854 & 0.9861\\
\hline
\end{tabular}}
\caption{effect of FC width in orientation accuracy}
\label{tab:tab1}
\end{table}

\subsection{Implicit Emergent Attention}
\begin{figure}
\begin{tabular}{c@{\hspace{1mm}}c@{\hspace{1mm}}c@{\hspace{1mm}}c}
\includegraphics[width = 0.11\textwidth]{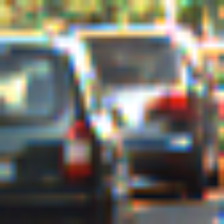}&
\includegraphics[width = 0.11\textwidth]{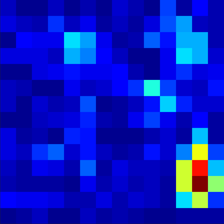}&
\includegraphics[width = 0.11\textwidth]{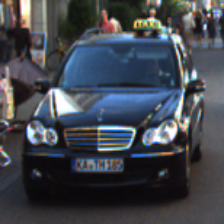}&
\includegraphics[width = 0.11\textwidth]{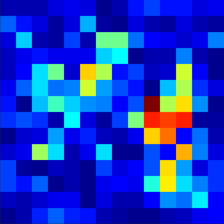}\\
\includegraphics[width = 0.11\textwidth]{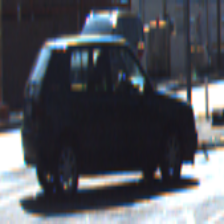}&
\includegraphics[width = 0.11\textwidth]{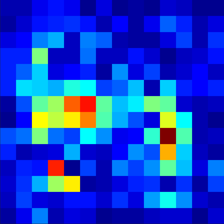}&
\includegraphics[width = 0.11\textwidth]{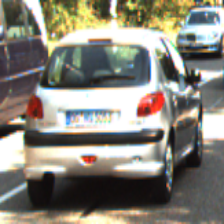}&
\includegraphics[width = 0.11\textwidth]{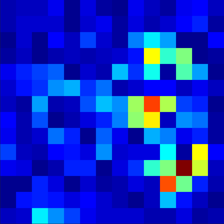}\\
\includegraphics[width = 0.11\textwidth]{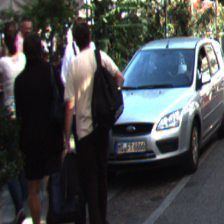}&
\includegraphics[width = 0.11\textwidth]{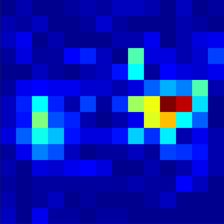}&
\includegraphics[width = 0.11\textwidth]{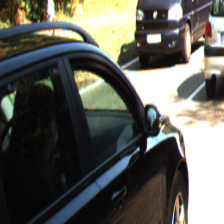}&
\includegraphics[width = 0.11\textwidth]{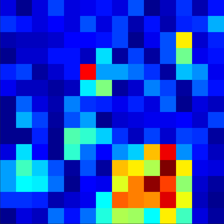}\\

\end{tabular}
\caption{Visualization of the learned attention of the model for orientation estimation. The heatmap shows the image areas that contribute to orientation estimation the most. The network attends to certain meaningful parts of the car such as tires, lights, and side mirrors.}
\label{fig:sliding_window}
\end{figure}

In this section, we visualize the parts of cars and bicycles that the network uses in order to estimate the object orientation accurately. Similar to \cite{Zhou_iclr15}, a small gray patch is slid around the image and for each location we record the difference between the estimated and the ground truth orientation. If occluding a specific part of the image by the patch causes a significantly different output, it means that the network attends to that part. 
Fig.~\ref{fig:sliding_window} shows such heatmaps of the output differences due to grayed out locations for several car detections. It appears that the network attends to distinct object parts such as tires, lights and side mirror for cars.
Our method seems to learn local features similar to keypoints used by other methods, without ever having seen explicitly labeled keypoint ground truth. Another advantage is that our network learns task-specific local features, while human-labeled keypoints are not necessarily the best ones for the task.

\ifarxiv
\subsection{Alternative Representation}
\label{sec:alternative_representation}
In this section we demonstrate the importance of choosing suitable regression parameters within our estimation framework. 
Here instead of object dimensions, we regress the location of the 3D box center projection in the image. This allows us to recover the camera ray towards the 3D box center. Any point on that ray can be described by a single parameter $\lambda$ which is the distance from the camera center. Given the projection of the center of the 3D box and the box orientation, our goal is to estimate $\lambda$ and the object dimensions: four unknowns for which we have four constraints between 2D box sides and 3D box corners. While the number of parameters to be regressed in this representation is less than those of the proposed method, this representation is more sensitive to regression errors. When there is no constraint on the physical dimension of the box, the optimization tries to satisfy the 2D detection box constraints even if the final dimensions are not plausible for the category of the object.

In order to evaluate the robustness of this representation, we take the ground truth 3D boxes and add realistic noise either to the orientation or to the location of the center of the 3D bounding box while keeping the enclosing 2D bounding box intact. The reason that we added noise was to simulate the parameter estimation errors. 3D boxes reconstructed using this formation satisfy the 2D-3D correspondences but have large box dimension errors as result of small errors in the orientation and box center estimates, as shown in Fig.~\ref{fig:failure}. This investigation supports our choice of 3D regression parameters. 

\begin{figure}
\begin{tabular}{c}                           
\includegraphics[width = 0.45\textwidth]{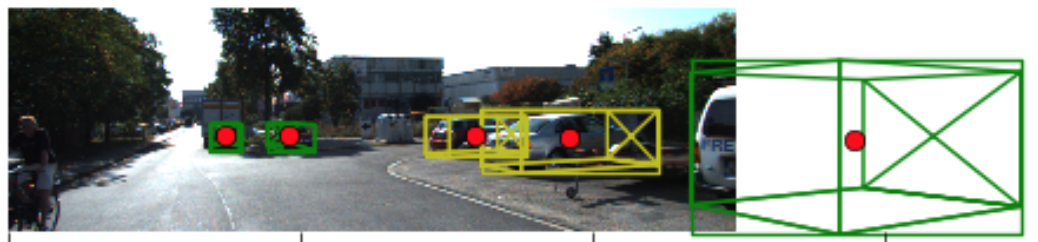}\\
\includegraphics[width = 0.45\textwidth]{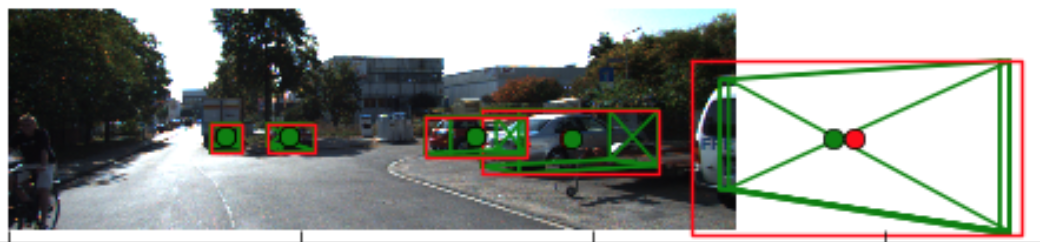}\\
\end{tabular}
\caption{Illustration of the sensitivity of the alternative representation that is estimating both dimensions and translation from geometric constraints. We added small amount of noise to the ground truth angles and tried to recover the ground truth box again. All other parameters are set to the ground truth values.}
\label{fig:failure}
\end{figure}
\fi

\section{Conclusions and Future Directions}
In this work, we show how to recover the 3D bounding boxes for known object categories from a single view. Using a novel MultiBin loss for orientation prediction and an effective choice 
of box dimensions as regression parameters, our method estimates stable and accurate posed 3D bounding boxes without additional 3D shape models, or sampling strategies with complex pre-processing pipelines. One future direction is to explore the benefits of augmenting the RGB image input in our method with a separate depth channel computed using stereo.  Another is to explore 3D box estimation in video, which requires using the temporal information effectively and can enable the prediction of future object position and velocity.

{\small
\bibliographystyle{ieee}
\balance\bibliography{egpaper_for_review}
}

\end{document}